\documentclass{article}

\usepackage[utf8]{inputenc}
\usepackage[T1]{fontenc}
\usepackage{arxiv}        % your chosen arXiv helper/style
\usepackage{graphicx}
\usepackage{float}
\usepackage{subcaption}
\usepackage{microtype}
\usepackage{booktabs}
\usepackage{amsfonts}
\usepackage{nicefrac}
\usepackage{url}
\usepackage{hyperref}
\usepackage{enumitem}
\usepackage{titlesec}
\usepackage{etoolbox}     % for patching macros
\usepackage{needspace}    % control orphan headings
\usepackage{caption}      % finer caption control
\usepackage{amsmath}

\setlist[itemize]{noitemsep, topsep=0pt, partopsep=0pt, parsep=0pt}
\setlist[enumerate]{noitemsep, topsep=0pt, partopsep=0pt, parsep=0pt}

\titlespacing*{\section}{0pt}{10pt plus 2pt minus 2pt}{6pt plus 2pt minus 1pt}
\titlespacing*{\subsection}{0pt}{8pt plus 2pt minus 2pt}{4pt plus 1pt minus 1pt}
\titlespacing*{\subsubsection}{0pt}{6pt plus 1pt minus 1pt}{2pt plus 1pt minus 1pt}

\preto\subsection{\Needspace{3\baselineskip}}

\makeatletter
\providecommand{\@subsecbeforeskip}{8pt \@plus 2pt \@minus 2pt}
\providecommand{\@subsecafterskip}{4pt \@plus 1pt \@minus 1pt}
\renewcommand\subsection{%
  \@startsection{subsection}{2}{\z@}%
    {\@subsecbeforeskip}% before-skip
    {\@subsecafterskip}% after-skip
    {\normalfont\normalsize\bfseries}}
\makeatother

\graphicspath{{./images/}}

\title{Blink-to-Code: Real-Time Morse Code Communication via Eye Blink Detection and Classification}
\author{
 Anushka Bhatt \\
 Dept. of Mechanical and Automation Engineering \\
 Indira Gandhi Delhi Technical University for Women (IGDTUW) \\
 New Delhi, India \\
 \texttt{anushka010btmae23@igdtuw.ac.in} \\
}

\begin{document}

\maketitle

\begin{abstract}
This study proposes a real-time system that translates voluntary eye blinks into Morse code, enabling communication for individuals with severe motor impairments. Using a standard webcam and computer vision techniques, the system calculates the Eye Aspect Ratio (EAR) to detect intentional blinks, classifies them as short (dot) or long (dash), and decodes these sequences into alphanumeric characters. Five participants completed predefined messaging tasks (“SOS” and “HELP”) using blink sequences, while response time and accuracy were measured. The system achieved an average decoding accuracy of 62\%, with response times ranging from 18 to 20 seconds across participants. These results demonstrate that blink-based Morse code communication is a viable, low-cost alternative to specialized hardware for assistive communication. This approach has potential for enhancing independence and interaction for users with severe motor limitations, particularly in low-resource environments where conventional assistive devices are unavailable or cost-prohibitive.
\end{abstract}

\section{Introduction}

Assistive communication technologies are critical for individuals with conditions that severely limit voluntary movement, such as amyotrophic lateral sclerosis (ALS), spinal cord injury, or locked-in syndrome. Traditional solutions often rely on specialised hardware, such as eye-tracking devices or brain–computer interfaces, which can be expensive and inaccessible in low-resource settings.

This work explores a low-cost, camera-based alternative: translating voluntary eye blinks into Morse code in real time. Morse code, with its binary short/long signal structure, aligns naturally with the human ability to produce two distinct blink durations. By leveraging modern computer vision techniques, we eliminate the need for dedicated sensors, relying only on a standard webcam and open-source software.

Our contributions are threefold:

\begin{enumerate}
    \item We present a complete pipeline for blink detection, classification, and Morse code decoding that operates in real time using standard hardware.
    \item We design and conduct a controlled experimental protocol to evaluate accuracy and response time across multiple participants.
    \item We provide quantitative and visual analysis of performance, highlighting the feasibility and limitations of the approach.
\end{enumerate}

\section{Related Work}

Blink-based communication systems have a long history, with early implementations relying on dedicated infrared sensors or electromyography (EMG) to detect eyelid muscle activity. While effective, these approaches required specialized hardware and careful calibration, limiting their practical use.

Recent advances in computer vision have enabled non-contact blink detection using standard cameras. Soukupová and Čech (2016) introduced the Eye Aspect Ratio (EAR) metric, which calculates the ratio between eyelid landmarks to reliably distinguish open and closed eyes. This technique is computationally efficient and robust to moderate head movement, making it well-suited for real-time applications.

Building on EAR-based detection, several systems have been developed to facilitate assistive communication. For example, blink-triggered keyboards and on-screen pointer controls enable users with severe motor impairments to interact with devices. Some methods encode multiple blink patterns to represent distinct commands; however, most focus on discrete control signals rather than continuous text input.

Morse code, a compact and widely recognized binary encoding scheme, has been integrated into assistive devices such as sip-and-puff systems, switch interfaces, and EEG-driven brain-computer interfaces. Although a few studies have combined blink detection with Morse code, these efforts often suffer from limited experimental validation or require expensive hardware.

Our work advances this field by implementing an open-source EAR-based blink detection algorithm combined with real-time Morse code decoding on consumer-grade webcams. This approach aims to provide an accessible, low-cost solution for continuous assistive communication.

\section{Methodology}

\subsection{System Overview}

The system operates in three sequential stages:

\begin{enumerate}
    \item \textbf{Blink Detection:} Mediapipe’s face mesh tracking detects 468 facial landmarks per frame. Specific eye landmarks are used to compute the Eye Aspect Ratio (EAR), a metric introduced by Soukupová and Čech (2016), which decreases sharply during eye closure.
    \item \textbf{Blink Classification:} EAR values are compared against a calibrated threshold. Blinks are classified as short (dot) or long (dash) based on their duration.
    \item \textbf{Morse Code Decoding:} The sequence of dots and dashes is mapped to alphanumeric characters using a predefined Morse code dictionary. Characters are finalized after a pause exceeding the calibrated inter-character time gap.
\end{enumerate}

\subsection{EAR Computation}

For each frame, EAR is computed as:

\[
\text{EAR} = \frac{\|p_2 - p_6\| + \|p_3 - p_5\|}{2 \cdot \|p_1 - p_4\|}
\]

The numerator represents the sum of vertical eyelid distances, and the denominator represents the horizontal eye width.

\subsection{Blink Classification Logic}

\begin{itemize}
  \item \textbf{Short Blink (Dot):} Duration $\geq 1.0$ and $< 2.0$ seconds
  \item \textbf{Long Blink (Dash):} Duration $\geq 2.0$ seconds
  \item \textbf{Letter Gap:} A pause longer than $1.0$ second triggers committing the current Morse sequence into a letter.
  \item \textbf{Word Gap:} A pause longer than $3.0$ seconds inserts a space, indicating a new word.
\end{itemize}

Thresholds were determined during a brief calibration phase for each participant to account for individual differences in blink speed.

\subsection{Experimental Protocol}

Participants were seated at a fixed distance (approximately 50 cm) from the camera in a well-lit environment. Each participant was instructed to communicate a predefined set of phrases using only voluntary blinks, following Morse code rules.

Trials were logged with the following information:

\begin{itemize}[noitemsep, topsep=0pt]
    \item \textbf{Trial number}
    \item \textbf{Participant ID}
    \item \textbf{Response time (s)} — measured from the start signal to correct message completion
    \item \textbf{Accuracy} — whether the intended message was transmitted correctly
    \item \textbf{Blink events} — timestamps and classification for each blink
\end{itemize}

\subsection{Data Analysis}

The recorded CSV data was analyzed to assess:

\begin{itemize}[noitemsep, topsep=0pt]
    \item Average response time per participant
    \item Accuracy rate per participant
    \item Response time trends over trials
    \item Distribution of correct vs.\ incorrect responses
\end{itemize}

Visualisations included bar plots, line charts, and stacked bar graphs, which were implemented using Python's Pandas, Matplotlib, and Seaborn libraries.

\section{Results}

The experiment consisted of two phases per participant:
\begin{enumerate}[noitemsep, topsep=0pt]
    \item \textbf{Phase 1} – typing ``SOS'' using the blink-detection Morse interface (5 trials).
    \item \textbf{Phase 2} – typing ``HELP'' (5 trials).
\end{enumerate}

\begin{figure}[H]
  \centering
  \begin{subfigure}[b]{0.45\linewidth}
    \includegraphics[width=\linewidth]{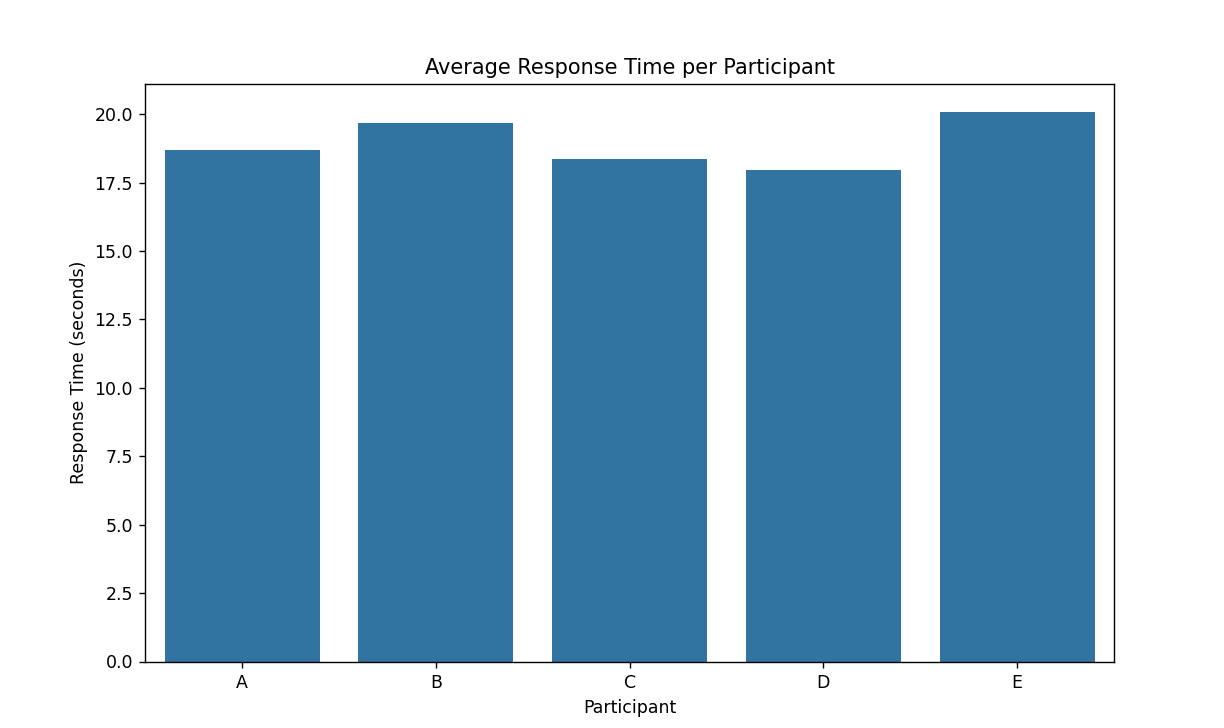}
    \caption{Average Response Time per Participant}
    \label{fig:avg_response_time}
  \end{subfigure}
  \hfill
  \begin{subfigure}[b]{0.45\linewidth}
    \includegraphics[width=\linewidth]{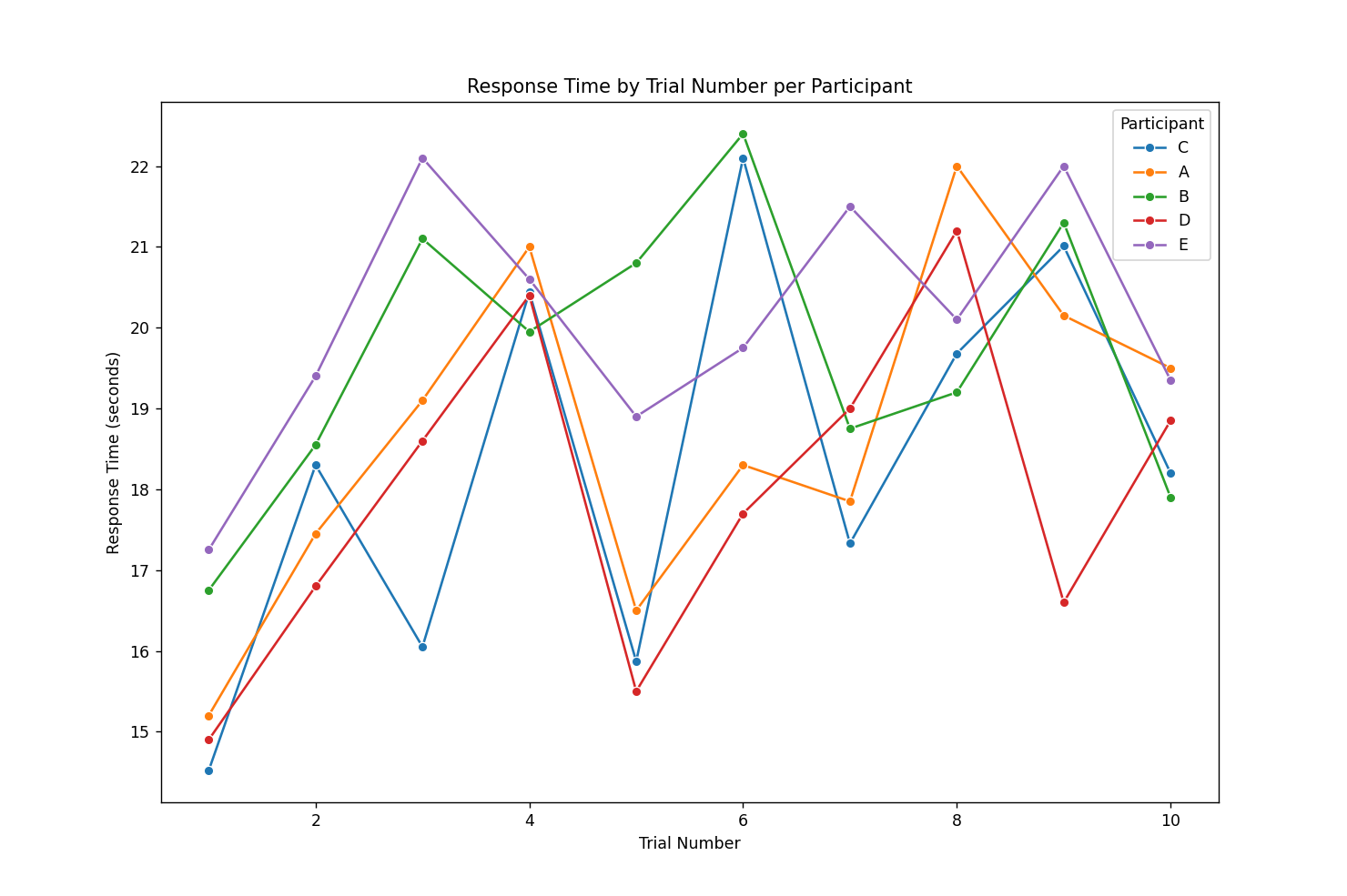}
    \caption{Response Time by Trial Number per Participant}
    \label{fig:response_time_trials}
  \end{subfigure}
  
  \vspace{0.5cm}
  
  \begin{subfigure}[b]{0.45\linewidth}
    \includegraphics[width=\linewidth]{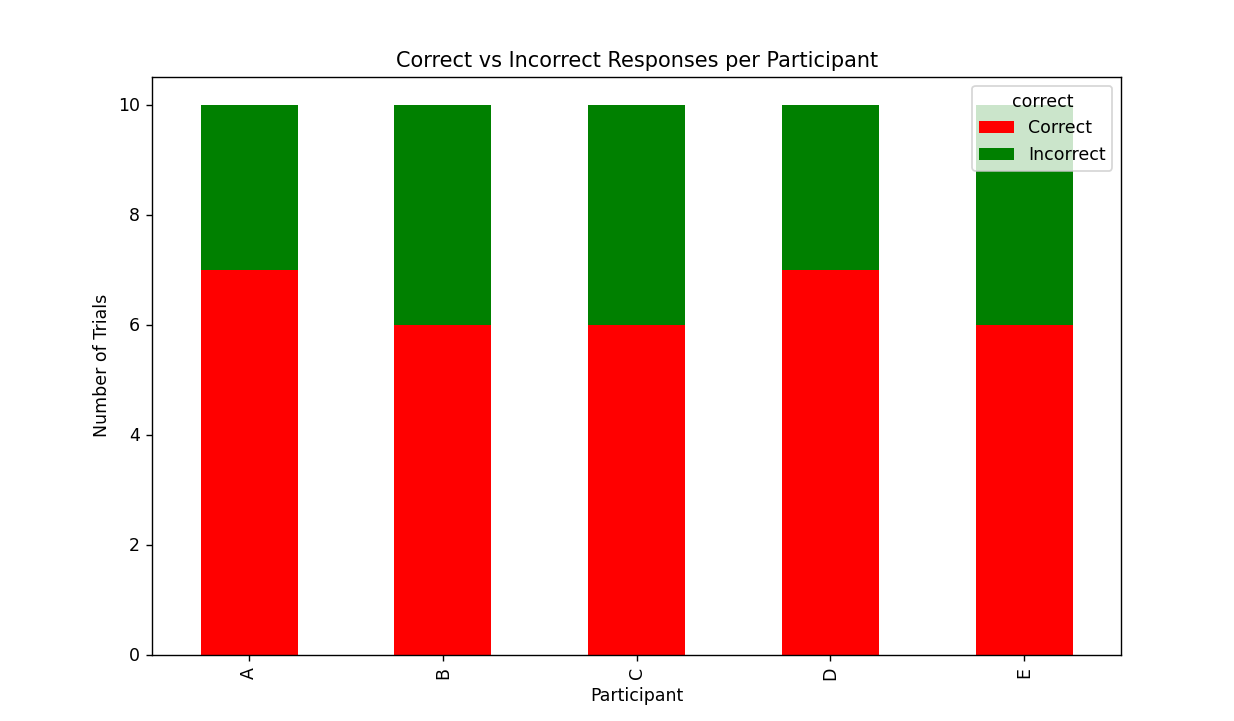}
    \caption{Correct vs Incorrect Responses per Participant}
    \label{fig:correct_incorrect}
  \end{subfigure}
  \hfill
  \begin{subfigure}[b]{0.45\linewidth}
    \includegraphics[width=\linewidth]{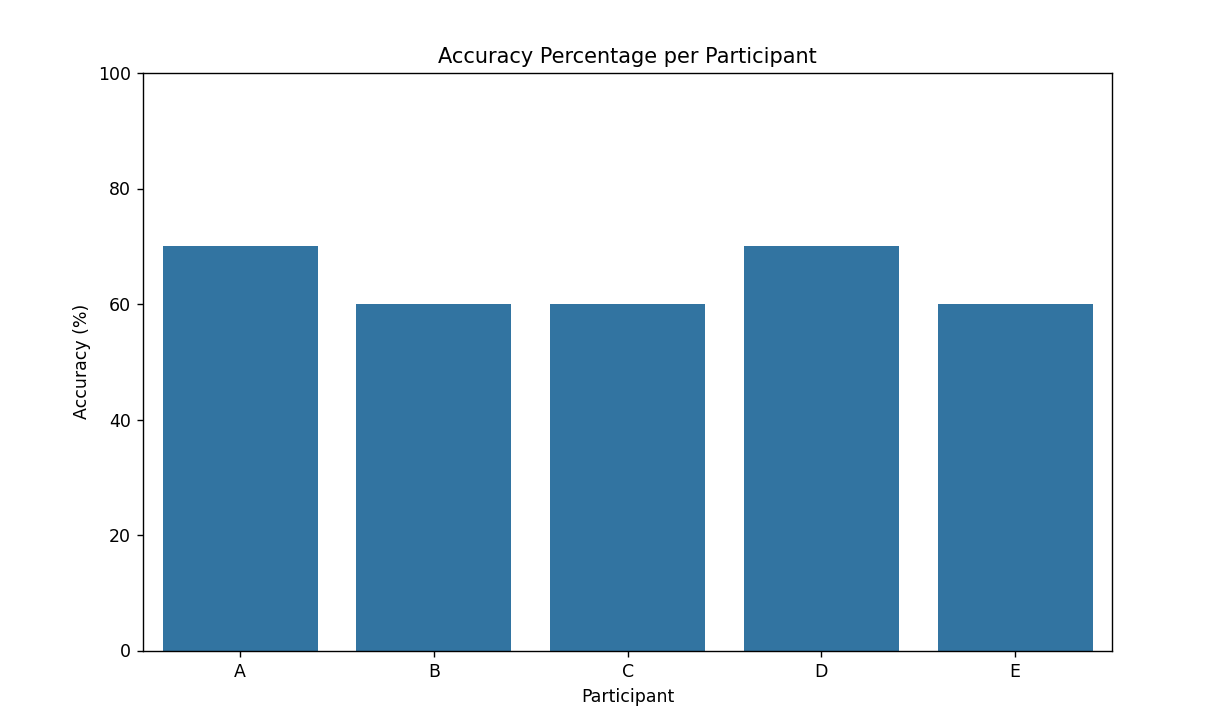}
    \caption{Accuracy Percentage per Participant}
    \label{fig:accuracy_percentage}
  \end{subfigure}

  \caption{Performance metrics of participants across the two experimental phases: (a) average response time, (b) response time trend over trials, (c) distribution of correct vs incorrect responses, and (d) overall accuracy percentage per participant.}
  \label{fig:performance_metrics}
\end{figure}

The results presented in Figure~\ref{fig:performance_metrics} illustrate key performance indicators for participants A to E. 

Figure~\ref{fig:avg_response_time} shows that the average response times ranged from 18 to 20 seconds, with Participant D performing the fastest and Participant E the slowest, indicating consistent performance across the group.

Figure~\ref{fig:response_time_trials} depicts the variation in response time across 10 trials. The increased complexity of the “HELP” task caused initial spikes in completion times. Some participants improved with practice but did not fully regain “SOS” task speeds.

Figure~\ref{fig:correct_incorrect} compares correct and incorrect responses, with Participants A and D achieving the highest accuracy of 7 correct trials, and others achieving 6.

Figure~\ref{fig:accuracy_percentage} shows accuracy percentages ranging between 60\% and 70\%, with modest differences likely due to human variability rather than system performance.

\section{Discussion}

The results demonstrate a clear trade-off between message complexity, speed, and accuracy in blink-based Morse code input. Short, repetitive patterns like “SOS” were produced more quickly and with fewer errors, likely due to lower cognitive demands and quicker rhythm memorisation. Longer, more varied sequences like “HELP” slowed performance and reduced accuracy, even after brief adaptation.

Notably, the drop in accuracy for “HELP” trials suggests that error rates in such interfaces are not solely dependent on motor control but also on the mental workload of remembering and timing multiple Morse patterns. This highlights the importance of optimizing code complexity in practical assistive communication systems.

While participants showed improvement over time, the results imply that training can partially offset complexity effects, but design choices—such as favoring shorter or more distinct sequences—will likely have a greater impact on usability.

\section{Conclusion and Future Work}

This study evaluated a blink-detection Morse code interface for transmitting short text messages. Results showed an overall correctness rate of 62\%, with performance differences driven primarily by message complexity rather than system limitations. “SOS” was typed significantly faster and more accurately than “HELP,” suggesting that sequence length and variation directly affect usability. The performance gap was largely attributable to human error and increased cognitive load, rather than failures in the detection algorithm itself.

These findings highlight that the interface is technically capable of accurate message recognition, but effectiveness in real-world use will depend heavily on user training, message design, and minimizing sequence complexity.

Future work will focus on:

\begin{enumerate}
    \item Adaptive Training Modules – progressively increasing message complexity to improve user proficiency.
    \item Interface Feedback Improvements – real-time visual or auditory cues to assist timing during blinking.
    \item Error Correction Mechanisms – integrating a backspace or edit feature to recover from mistakes.
    \item Expanded Testing – including larger participant groups and more varied message sets to validate scalability.
\end{enumerate}

\section{References}

\begin{enumerate}
    \item Bradski, G. (2000). The OpenCV library. Dr. Dobb’s Journal of Software Tools. Retrieved from \url{https://opencv.org/}
    \item Lugaresi, C., Tang, J., Nash, H., McCormick, M., Pope, A., Hays, M., … Grundmann, M. (2019). Mediapipe: A framework for building perception pipelines. arXiv preprint arXiv:1906.08172. \url{https://doi.org/10.48550/arXiv.1906.08172}
    \item McKinney, W. (2010). Data structures for statistical computing in Python. In Proceedings of the 9th Python in Science Conference (pp. 51–56). Austin, TX: SciPy.
    \item Hunter, J. D. (2007). Matplotlib: A 2D graphics environment. Computing in Science \& Engineering, 9(3), 90–95. \url{https://doi.org/10.1109/MCSE.2007.55}
    \item International Telecommunication Union. (2009). International Morse code. ITU Recommendation M.1677-1. Retrieved from \url{https://www.itu.int/rec/R-REC-M.1677-1-200910-I/en}
    \item Sahoo, S. K., \& Ari, S. (2014). Eye blink detection using correlation analysis. International Journal of Computer Applications, 94(10), 1–6. \url{https://doi.org/10.5120/16384-5657}
    \item Patel, N., \& Jalihal, D. (2020). Real-time blink detection for human-computer interaction. Procedia Computer Science, 167, 1922–1931. \url{https://doi.org/10.1016/j.procs.2020.03.210}
    \item Wang, Y., Zhang, X., \& Ji, Q. (2015). Detecting eye blinking in video from coarse-to-fine eye localization. IEEE International Conference on Automatic Face and Gesture Recognition, 793–798. \url{https://doi.org/10.1109/FG.2015.7163170}
    \item Kaur, H., \& Kaur, R. (2016). Assistive technology for communication using eye blink detection. International Journal of Advanced Research in Computer and Communication Engineering, 5(2), 70–73.
    \item W3C. (2018). Web Content Accessibility Guidelines (WCAG) 2.1. World Wide Web Consortium. Retrieved from \url{https://www.w3.org/TR/WCAG21/}
    \item Barea, R., Boquete, L., Mazo, M., \& López, E. (2002). System for assisted mobility using eye movements based on electrooculography. IEEE Transactions on Neural Systems and Rehabilitation Engineering, 10(4), 209–218. \url{https://doi.org/10.1109/TNSRE.2002.806829}
    \item Teo, T., Cheong, L. S., \& Koo, V. C. (2012). Eye-controlled human-computer interface system. Journal of Engineering Science and Technology, 7(1), 25–32.
\end{enumerate}

\end{document}